\documentclass[conference]{IEEEtran}
\IEEEoverridecommandlockouts
\usepackage{cite}
\usepackage{subcaption}
\usepackage{float}
\usepackage{amsmath,amssymb,amsfonts}
\usepackage{algorithmic}
\usepackage{graphicx}
\usepackage{textcomp}
\usepackage{xcolor}
\usepackage{orcidlink}
\usepackage{hyperref}
\def\BibTeX{{\rm B\kern-.05em{\sc i\kern-.025em b}\kern-.08em
    T\kern-.1667em\lower.7ex\hbox{E}\kern-.125emX}}

\title{TGO-III: Semantic Geometry Observatory}
\author{
\IEEEauthorblockN{Kaustubh Kapil\orcidlink{0009-0000-4918-8452} and Kishor P. Upla\orcidlink{0000-0001-6306-0682}}
\IEEEauthorblockA{
Department of Electronics Engineering\\
Sardar Vallabhai National Institute of Technology (SVNIT), Surat, India\\
u24ec049@eced.svnit.ac.in, kishorupla@gmail.com
}
}

\begin{document}

\maketitle

\begin{abstract}
With the widespread adoption of Vision Transformers in modern AI, the need to analyze their inherent representational behavior has become increasingly important. While most existing studies emphasize token geometries and training dynamics, the evolution of representational covariance structures and class-level geometric organization remains comparatively underexplored. In this work, we investigate semantic geometry and class separability as representations evolve across the layers of ViT-Small/16 through TGO-III: Semantic Geometry Observatory. It is a framework designed to analyze the emergence of semantic organization, feature evolution, and class-wise representation geometry throughout training. The framework employs multiple complementary observatories, including Linear Probe Accuracy, Fisher Ratio, Class Centroid Distances, Local Intrinsic Dimension, and Local PCA Rank, to quantify the progressive evolution of discriminative representations. Our analysis reveals that class representations become progressively more linearly separable, Fisher discriminability increases, class centroids move farther apart, and local representation manifolds exhibit structured class-dependent geometric complexity. These observations provide empirical evidence supporting the Semantic Expansion Hypothesis, suggesting that the manifold expansion observed in previous observatories is accompanied by the progressive organization of representations into increasingly discriminative semantic structures. Collectively, TGO-III extends the Transformer Geometry Observatory framework by establishing a direct connection between manifold geometry, covariance evolution, and semantic organization during Transformer training.
\end{abstract}

\section{Introduction}
The remarkable success of Vision Transformers (ViTs) has fundamentally reshaped modern computer vision, achieving state-of-the-art performance across image classification, object detection, semantic segmentation, and numerous downstream visual understanding tasks. Despite their empirical success, the internal representational dynamics governing how Transformers progressively organize visual information during training remain only partially understood. Existing studies have largely concentrated on attention mechanisms, token interactions, optimization dynamics, and representational similarity, providing valuable insights into how information propagates across Transformer layers. However, comparatively less attention has been devoted to understanding how semantic structure emerges within the learned representation space itself. With recent improvements in analyzing representation space, it has been shown that there is significant geometrical change in Transformer representations during the training process. Methods like spectral analysis, eigenspectrum changes, covariance analysis, representational similarity, and intrinsic dimensionality have been found to show that the feature space is being regularly changed due to optimization. Though these methods succeed in explaining the geometrical evolution of representations, they are mostly able to describe global properties of representation space.

The Transformer Geometry Observatory (TGO) framework was introduced to systematically investigate these representational dynamics from complementary geometric perspectives. \textbf{TGO-I: Spectral Geometry Observatory}~\cite{tgoi} demonstrated that the covariance spectrum progressively expands during training, accompanied by increasing Effective Rank, Stable Rank, and Spectral Entropy, suggesting that the network continuously explores additional representational directions. Subsequently, \textbf{TGO-II: Representation Geometry Observatory}~\cite{tgoii} revealed that the intrinsic dimensionality of Transformer representations increases throughout optimization while representational similarity between layers progressively decreases. These observations motivated the Manifold Expansion Hypothesis, proposing that the observed covariance expansion arises from the progressive expansion of the underlying representation manifold rather than token diversification alone. In order to solidify these claims and hypotheses, we introduce \textbf{TGO-III: Semantic Geometry Observatory}. a framework designed to investigate the emergence of semantic organization during Transformer training. Rather than studying global covariance or manifold properties alone, TGO-III analyzes how class representations evolve throughout optimization using complementary measurements including Linear Probe Accuracy, Fisher Ratio, Class Centroid Distances, Local Intrinsic Dimensionality, and Local PCA Rank. Collectively, these observatories quantify the evolution of class separability, local manifold complexity, and discriminative organization within the learned representation space. The key contributions of TGO-III can be summarized as follows:
\begin{itemize}
    \item This study performs a detailed analysis on class separability and the geometric evolution of representational covariances; with a major focus on class-level geometric organization
    \item Providing important insights on \emph{semantic hypotheses} mentioned through previous TGOs~\cite{tgoi,tgoii}
    \item Discussing fundamental semantic properties such as increasing inter-class distances, improving Fisher discriminability, and structured class-dependent complexity in local representation geometry.
\end{itemize}

\begin{figure*}[t]
    \centering
    \includegraphics[width=1.0\linewidth]{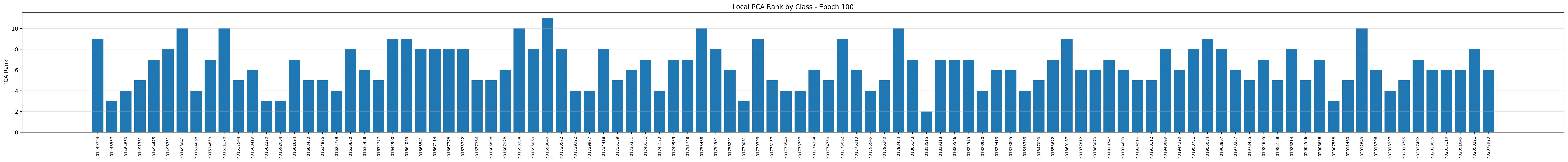}
    \caption{The final Local PCA Rank by class showing how many explorable dimensions each class needs}
    \label{fig:local_pca_final}
\end{figure*}

\subsection{Research Questions}
\begin{enumerate}

\item \textbf{RQ1: Does semantic organization progressively emerge throughout Transformer training?}

If the manifold expansion observed in previous observatories reflects the acquisition of increasingly meaningful representations, one would expect class separability to improve continuously throughout optimization. This question is investigated through Linear Probe Accuracy, Fisher Ratio, Class Centroid Distances, Local Intrinsic Dimension and Local PCA Rank.

\item \textbf{RQ2: Can the semantic organization of class representations explain the covariance expansion previously observed in TGO-I?}

TGO-I demonstrated a continuous increase in covariance rank and spectral complexity without identifying the representational origin of these additional covariance directions. This question investigates whether the progressive organization and separation of semantic classes provides a plausible geometric explanation for the observed covariance expansion.

\item \textbf{RQ3: How does local class geometry evolve during Transformer training?}

While previous observatories primarily characterized global properties of the representation manifold, little is known about the evolution of individual semantic classes. This question investigates how the intrinsic dimensionality and local geometric complexity of individual class manifolds evolve throughout optimization.

\item \textbf{RQ4: Is discriminative representation learning uniformly distributed across Transformer depth?}

Previous observatories suggested the existence of a transition zone separating qualitatively different stages of representation processing. This question investigates whether semantic discrimination emerges uniformly across all Transformer blocks or whether specific layers exhibit significantly stronger class specialization, thereby providing additional evidence regarding the Transition Zone Hypothesis.

\end{enumerate}
The complete implementation of TGO-III, trained checkpoints, observatory pipelines,
and generated artifacts are publicly available at:
\href{https://github.com/KaustubhKapil/Transformer_Geometry_Observatory_Part-3}{GitHub Repository}.

\begin{figure}[t]
    \centering
    \includegraphics[width=\columnwidth]{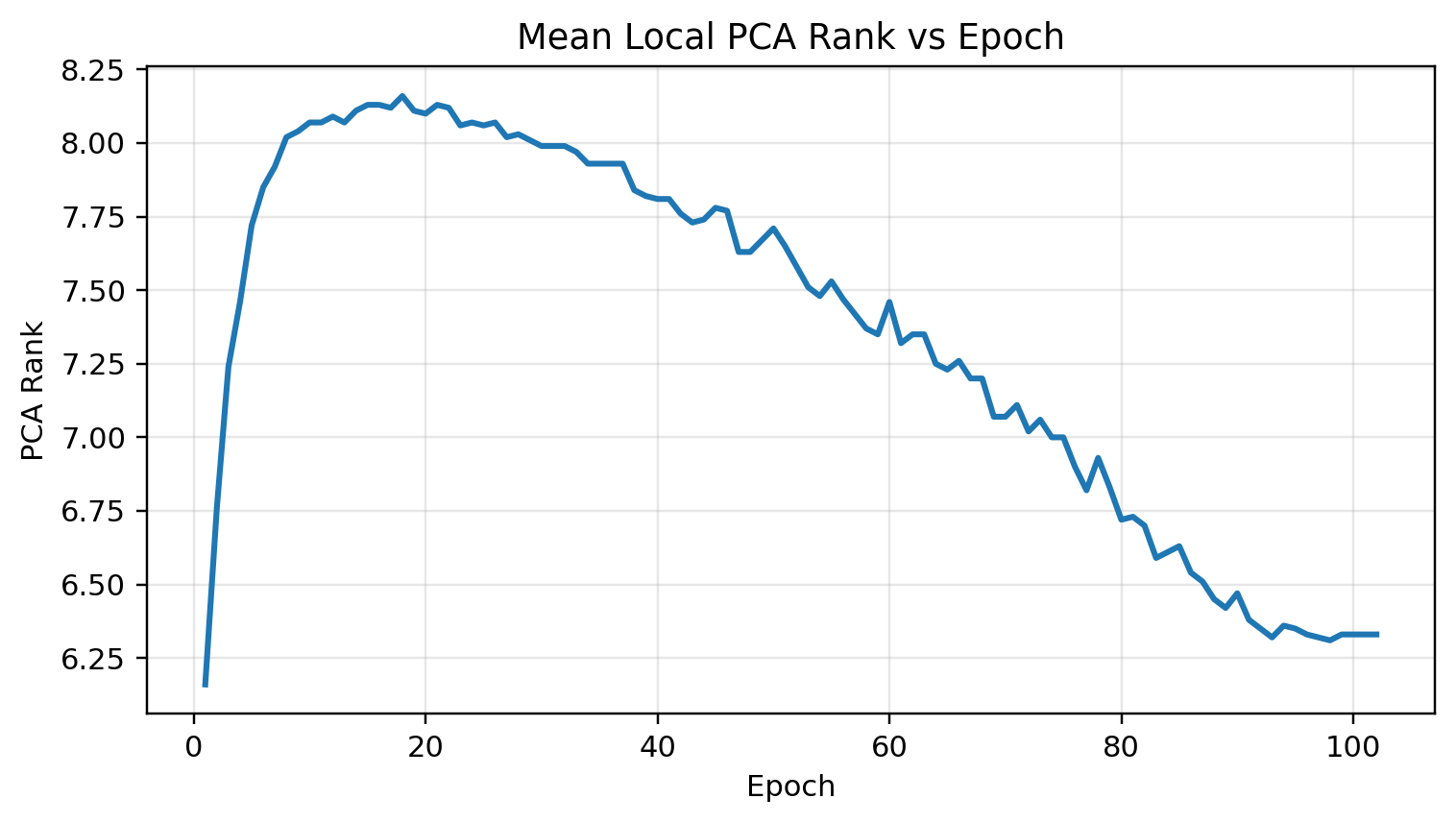}
    \caption{Evolution of the mean Local PCA Rank throughout Transformer training. The average Local PCA Rank initially increases during the early stages of optimization before progressively decreasing as training converges, indicating that semantic class manifolds become increasingly concentrated within lower-dimensional local subspaces.}
    \label{fig:local_pca_curve}
\end{figure}

\section{Methodology and Semantic Metrics}
\label{sec:frame}
The Semantic Geometry Observatory is designed to investigate how Transformer representations progressively evolve into semantically organized representations throughout training. The central concept underlying this framework is \emph{Semantic Expansion}, defined as the progressive geometric organization of the representation manifold, whereby newly explored representational directions are increasingly allocated toward encoding discriminative semantic concepts, resulting in improved class separability and richer class-level representation geometry throughout training. Unlike previous observatories, which primarily examined global spectral and representational properties, TGO-III focuses on the geometric organization of individual semantic classes. Specifically, the framework investigates how ImageNet-100~\cite{imagenet} class representations evolve across the layers of a Vision Transformer, how their local manifolds change throughout optimization, and how progressively concentrated and discriminative these class representations become during training. With the help of the following analyses tools TGO-III can make concrete conclusions.

\subsection{Linear Probe Accuracy}

Linear probing is a standard technique for quantifying the linear separability of learned representations. Given the feature representation
\begin{equation}
\mathbf{X}=\{\mathbf{x}_1,\mathbf{x}_2,\ldots,\mathbf{x}_N\},\qquad
\mathbf{x}_i\in\mathbb{R}^{D},
\end{equation}
together with the corresponding class labels
\begin{equation}
\mathbf{y}=\{y_1,y_2,\ldots,y_N\},
\end{equation}
a linear classifier is trained while keeping the Transformer weights fixed. Specifically, a linear decision function

\begin{equation}
f(\mathbf{x})=\mathbf{W}\mathbf{x}+\mathbf{b}
\end{equation}

is optimized using the extracted representations.

The performance of the probe is evaluated on a held-out validation set using classification accuracy

\begin{equation}
\mathrm{Accuracy}
=
\frac{1}{N}
\sum_{i=1}^{N}
\mathbf{1}
\left(
\hat{y}_i=y_i
\right),
\end{equation}
where $\hat{y}_i$ denotes the predicted label.

Unlike end-to-end classification accuracy, Linear Probe Accuracy measures the amount of class-discriminative information already encoded within the representation itself. Consequently, improvements in probe accuracy indicate increasing linear separability of the learned feature space without further optimization of the backbone network.

\begin{figure*}[t]
    \centering

    \subfloat[Fisher Ratio throughout training.\label{fig:fisher_curve}]{
        \includegraphics[width=0.48\textwidth]{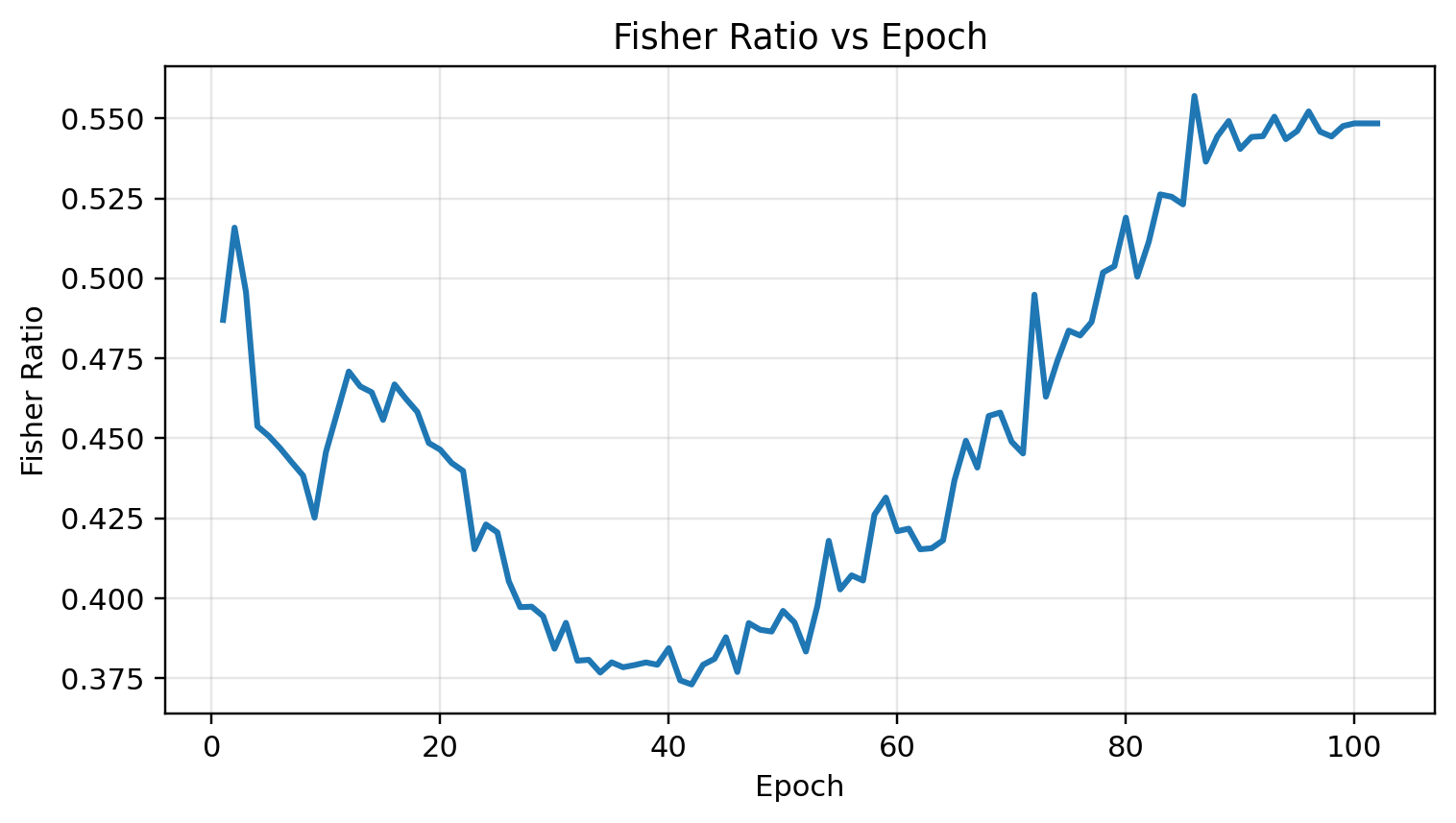}
    }
    \hfill
    \subfloat[Layer-wise Fisher Ratio at Epoch 100.\label{fig:fisher_layer_final}]{
        \includegraphics[width=0.48\textwidth]{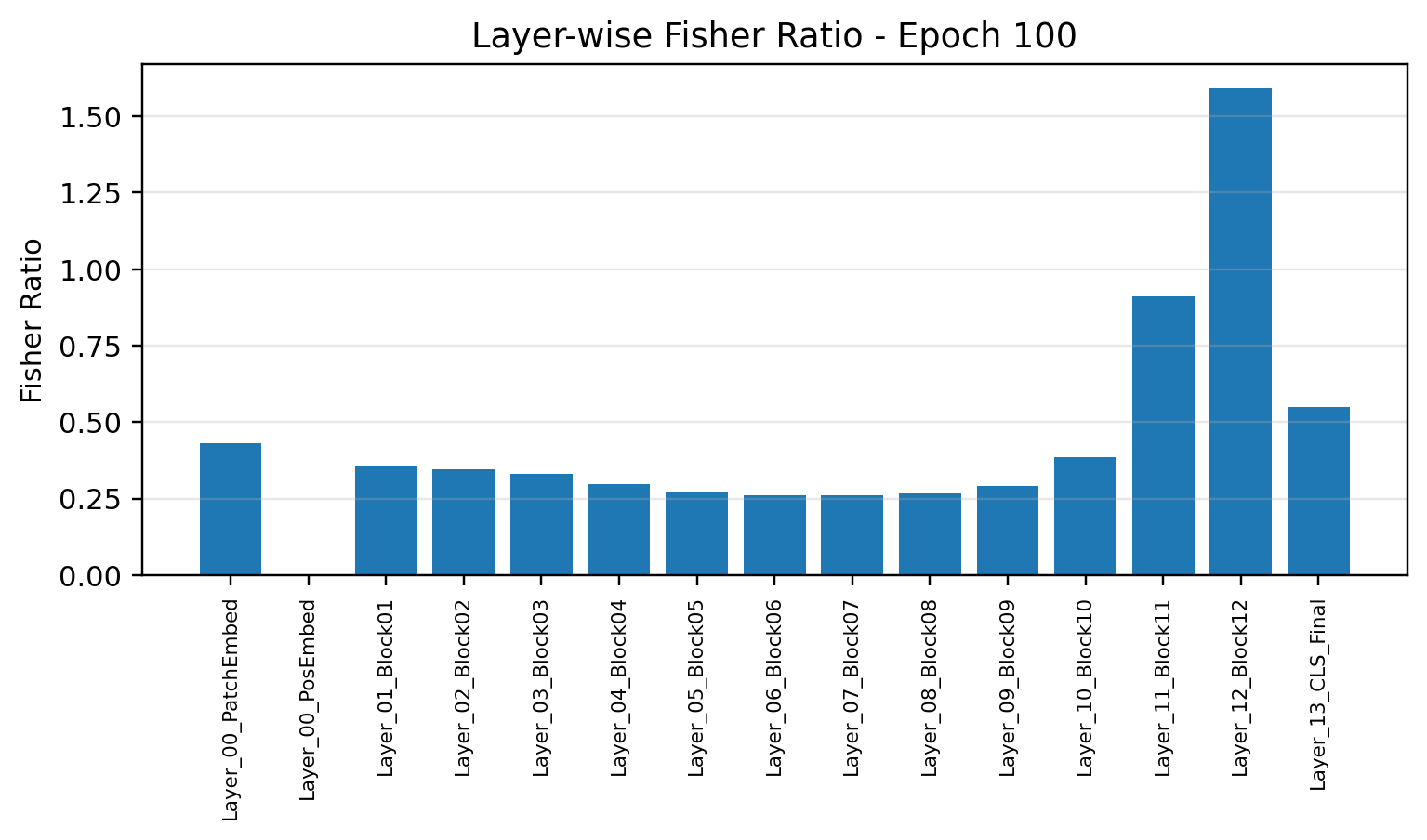}
    }

    \caption{Evolution of Fisher Ratio throughout Transformer training. The left figure illustrates the evolution of the mean Fisher Ratio across training epochs, while the right figure summarizes the Fisher Ratio of each Transformer layer at convergence (Epoch 100). Although the mean Fisher Ratio exhibits a two-stage evolution, decreasing during the initial stages of optimization before steadily increasing toward convergence, layer-wise analysis reveals that the strongest discriminative specialization is concentrated within the final Transformer blocks, with Block~12 exhibiting the highest Fisher Ratio.}
    \label{fig:fisher_results}
\end{figure*}

\subsection{Fisher Ratio}

To quantify class separability from a statistical perspective, TGO-III employs the Fisher Ratio, which measures the ratio between inter-class variance and intra-class variance.

Let the global feature mean be

\begin{equation}
\boldsymbol{\mu}
=
\frac{1}{N}
\sum_{i=1}^{N}
\mathbf{x}_i,
\end{equation}

and the centroid of class $c$

\begin{equation}
\boldsymbol{\mu}_c
=
\frac{1}{N_c}
\sum_{i:y_i=c}
\mathbf{x}_i.
\end{equation}

The within-class scatter is defined as

\begin{equation}
S_W
=
\sum_c
\sum_{i:y_i=c}
\|
\mathbf{x}_i-\boldsymbol{\mu}_c
\|^2,
\end{equation}

while the between-class scatter is

\begin{equation}
S_B
=
\sum_c
N_c
\|
\boldsymbol{\mu}_c-\boldsymbol{\mu}
\|^2.
\end{equation}

The Fisher Ratio is therefore

\begin{equation}
F
=
\frac{S_B}
{S_W+\varepsilon},
\end{equation}

where $\varepsilon$ is a small numerical constant for stability.

The Fisher Ratio jointly captures increasing inter-class separation and decreasing intra-class variability, making it an effective measure of discriminative representation quality.

\subsection{Class Centroid Distance}

The geometric organization of semantic classes is analyzed through pairwise distances between class centroids.

For each semantic class $c$, the centroid is computed as

\begin{equation}
\boldsymbol{\mu}_c
=
\frac{1}{N_c}
\sum_{i:y_i=c}
\mathbf{x}_i.
\end{equation}

The pairwise Euclidean distance between two classes is

\begin{equation}
D_{ij}
=
\|
\boldsymbol{\mu}_i-
\boldsymbol{\mu}_j
\|_2.
\end{equation}

The complete centroid distance matrix

\begin{equation}
\mathbf{D}
=
[D_{ij}]
\end{equation}

characterizes the global geometric arrangement of semantic classes within the representation space.

For quantitative analysis, the mean centroid distance is computed as

\begin{equation}
\bar{D}
=
\frac{2}
{C(C-1)}
\sum_{i<j}
D_{ij},
\end{equation}

where $C$ denotes the total number of semantic classes.

\subsection{Local PCA Rank}

While centroid distances characterize global semantic organization, Local PCA Rank measures the intrinsic geometric complexity of individual semantic classes.

For each class $c$, let

\begin{equation}
\mathbf{X}_c
=
\{\mathbf{x}_i
\mid
y_i=c
\}
\end{equation}

denote all feature vectors belonging to that class.

Principal Component Analysis is performed on the centered class representation

\begin{equation}
\tilde{\mathbf{X}}_c
=
\mathbf{X}_c-
\boldsymbol{\mu}_c.
\end{equation}

The covariance matrix is

\begin{equation}
\Sigma_c
=
\frac{1}{N_c-1}
\tilde{\mathbf{X}}_c^T
\tilde{\mathbf{X}}_c.
\end{equation}

Let

\begin{equation}
\lambda_1\ge
\lambda_2\ge
\cdots
\ge
\lambda_D
\end{equation}

be the eigenvalues of $\Sigma_c$.

The cumulative explained variance is

\begin{equation}
E(k)
=
\frac{
\sum_{i=1}^{k}
\lambda_i
}{
\sum_{i=1}^{D}
\lambda_i
}.
\end{equation}

The Local PCA Rank is defined as the minimum number of principal components satisfying

\begin{equation}
E(k)
\ge
\tau,
\end{equation}

where $\tau=0.95$ throughout this work.

Consequently, Local PCA Rank estimates the effective dimensionality required to represent the local geometry of an individual semantic class.

\section{Experiments}
This section dives into how the analyses were performed, describing hardware, presets and backbone details. A ViT-Small/16 model~\cite{vit} was trained on the ImageNet-100~\cite{imagenet}dataset for 100 epochs using a NVIDIA Quadro RTX 6000 GPU. To ensure consistent observability, all representational measurements were performed on a fixed analysis subset of 1000 validation images throughout training. TGO-III starts with considering input embedding matrix represented by $X_l$.
\begin{equation}
\mathbf{X}_l \in \mathbb{R}^{(B \times T)\times D},
\end{equation}
Since we analyze feature matrices and token matrices, the covariance matrix is considered, computed as
\begin{equation}
\mathbf{C}^{feat}_{l}
=
\frac{1}{N-1}
\left(
\mathbf{X}_{l}
-
\mathbf{1}\boldsymbol{\mu}_{l}^{T}
\right)^{T}
\left(
\mathbf{X}_{l}
-
\mathbf{1}\boldsymbol{\mu}_{l}^{T}
\right),
\end{equation}
where 
\begin{equation}
N = B \times T
\end{equation}
and
\begin{equation}
\boldsymbol{\mu}_{l}
=
\frac{1}{N}
\sum_{i=1}^{N}
\mathbf{X}_{l}^{(i)}
\end{equation}
where $\boldsymbol{\mu}_l$ denotes the sample mean vector computed across all representation vectors at layer $l$. The normalization factor $N-1$ corresponds to Bessel's correction, yielding an unbiased estimate of the sample covariance matrix. 

Similarly TGO-III computes the token covariance matrix as
\begin{equation}
\mathbf{C}^{token}_{l}
=
\frac{1}{D-1}
\left(
\mathbf{X}_{l}
-
\bar{\mathbf{X}}_{l}
\right)
\left(
\mathbf{X}_{l}
-
\bar{\mathbf{X}}_{l}
\right)^{T},
\end{equation}
where
\begin{equation}
\bar{\mathbf{X}}_{l}
=
\frac{1}{D}
\sum_{j=1}^{D}
\mathbf{X}_{l}^{(:,j)}
\end{equation}
denotes the token-wise mean representation. The token covariance matrix captures pairwise relationships between tokens and provides insight into token coupling and interaction structure throughout training. It is important to note that the feature covariance matrix and token covariance matrix are related through the same underlying representation matrix. Both the covariance matrices have the same non-zero singular values and eigenvalues, but the eigenvectors differ. The above relation makes it possible for us to consider feature space and token space analysis as two perspectives on the same geometrical representation.

\section{Findings}
\label{sec:find}

In this section, we investigate the semantic evolution of Transformer representations using the proposed observatories. Unlike previous TGOs, which primarily focused on global covariance structure and representational geometry, TGO-III analyzes the discriminative organization of semantic classes as training progresses. The following observations are presented independently of the hypotheses discussed in Section~\ref{sec:hyp}.

\subsection{Linear Probe Accuracy}

\begin{figure*}
    \centering

    \subfloat[Layer-wise Probe Accuracy throughout training.\label{fig:probe_layer_curve}]{
        \includegraphics[width=\textwidth]{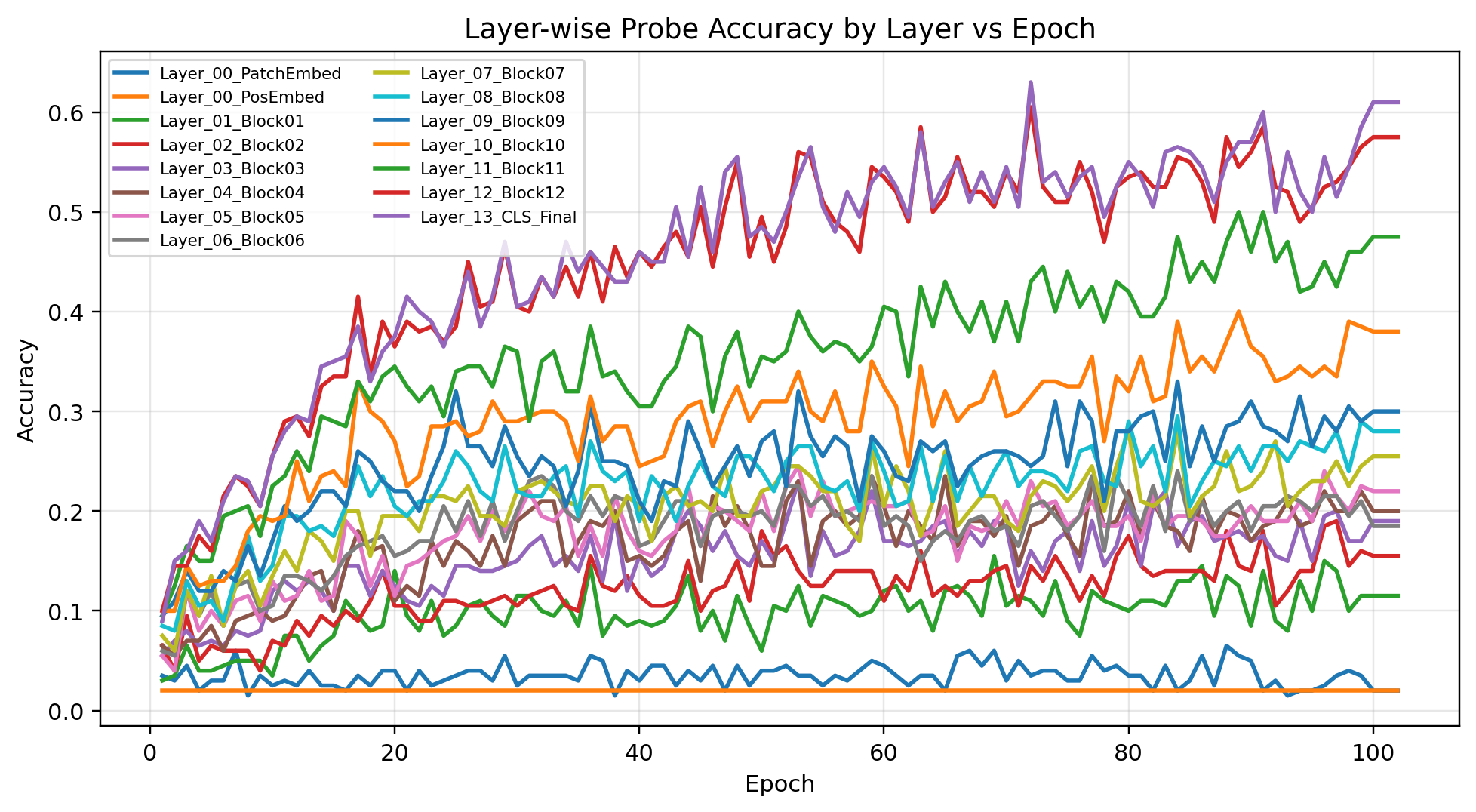}
    }

    \vspace{0.4cm}

        \subfloat[Layer-wise Probe Accuracy at Epoch 100.\label{fig:probe_layer_final}]{
        \includegraphics[width=0.48\textwidth]{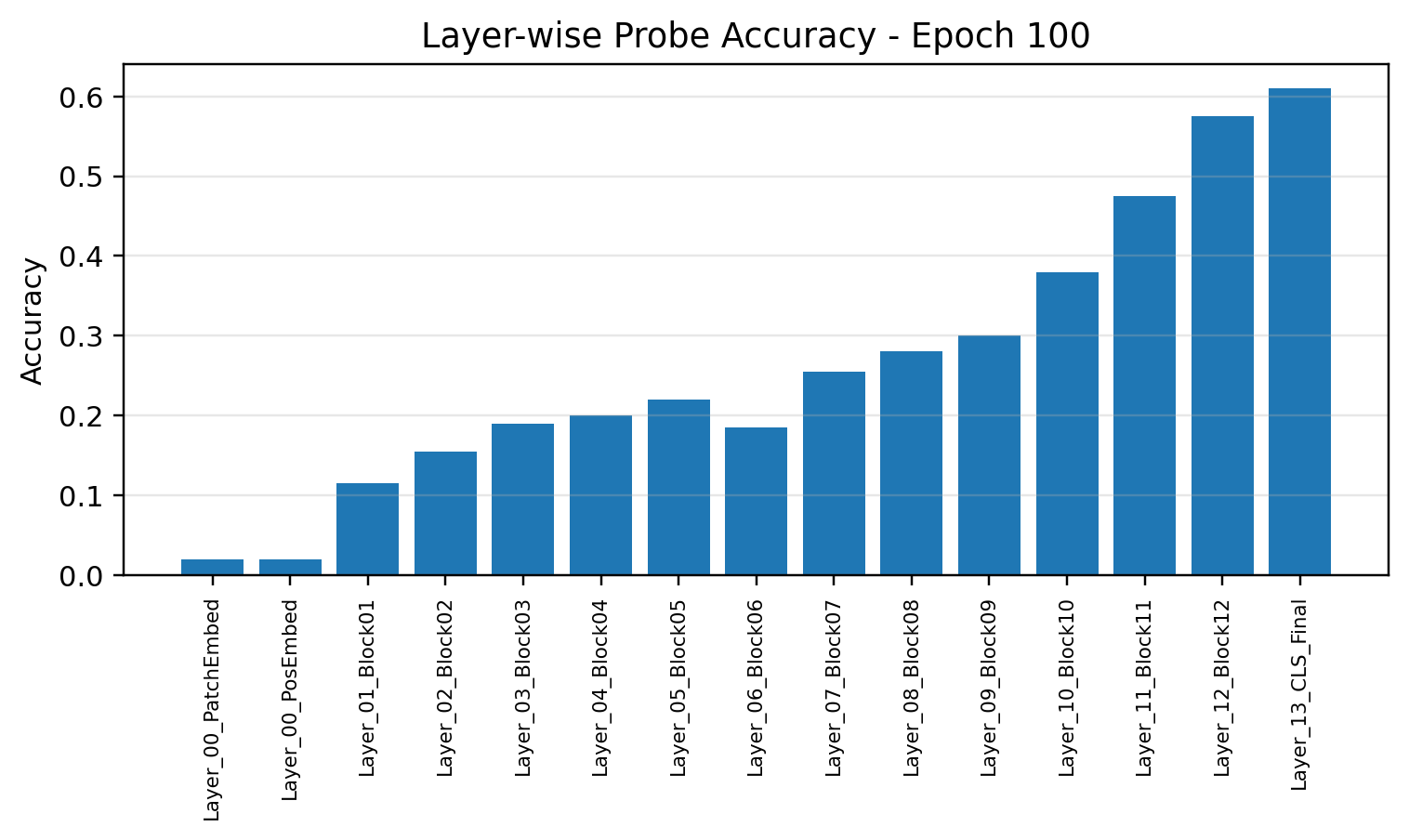}
    }
    \hfill
    \subfloat[Mean Probe Accuracy throughout training.\label{fig:probe_curve}]{
        \includegraphics[width=0.48\textwidth]{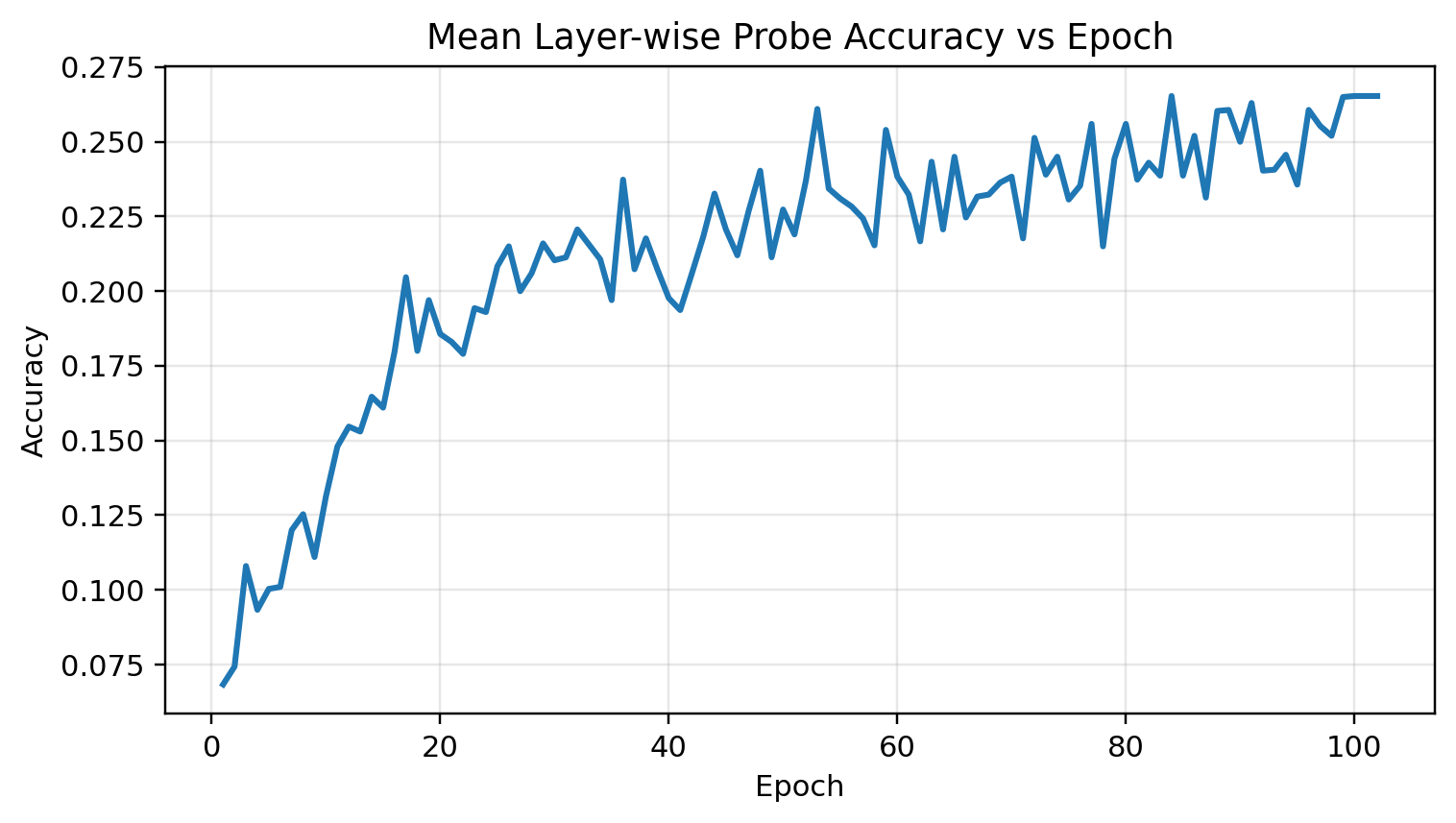}
    }

    \caption{Evolution of Linear Probe Accuracy throughout Transformer training. The top figure illustrates the layer-wise evolution of probe accuracy across all Transformer blocks throughout optimization. The bottom-right figure presents the evolution of the mean probe accuracy, while the bottom-left figure summarizes the probe accuracy of each Transformer layer at convergence (Epoch 100). Probe accuracy increases consistently throughout training, with the strongest linear separability emerging within the final Transformer blocks and reaching its maximum at the final CLS representation.}
    \label{fig:probe_results}
\end{figure*}

Linear probes are employed to measure the linear separability of semantic classes throughout training. Figures~\ref{fig:probe_layer_curve} and~\ref{fig:probe_layer_final} present the layer-wise probe accuracies during training and at the final epoch, respectively, while Figure~\ref{fig:probe_curve} illustrates the evolution of the average probe accuracy.

As training progresses, the average probe accuracy increases consistently, indicating that the learned representations become progressively easier to classify using a simple linear classifier. This suggests that semantic information becomes increasingly organized within the representation space throughout optimization. Layer-wise analysis reveals a clear hierarchical trend. Early Transformer blocks exhibit relatively poor probe accuracies, whereas the later Transformer blocks achieve substantially higher accuracies. The final classification representation produces the highest probe accuracy among all extracted representations, demonstrating that semantic information accumulates progressively across the Transformer depth. A noticeable acceleration in probe accuracy is observed after approximately the tenth Transformer block, indicating that the majority of semantic discrimination is developed within the later stages of the encoder.

\subsection{Fisher Ratio}

The Fisher Ratio evaluates the ratio between inter-class scatter and intra-class scatter, thereby quantifying the discriminative quality of the learned representations. Figure~\ref{fig:fisher_curve} illustrates the evolution of the average Fisher Ratio during training, while Figure~\ref{fig:fisher_layer_final} present the corresponding layer-wise evolution. The average Fisher Ratio exhibits two distinct phases throughout optimization. During the initial half of training, the Fisher Ratio gradually decreases before reaching a minimum around the middle of optimization. Subsequently, the Fisher Ratio increases steadily until convergence, ultimately exceeding its initial value. This behaviour suggests that the early stages of optimization primarily reorganize the representation manifold before progressively increasing class discrimination during later training. Layer-wise Fisher analysis reveals a highly non-uniform distribution across the Transformer. Most intermediate blocks exhibit relatively similar Fisher Ratios, whereas the final Transformer blocks display a dramatic increase in discriminative power. In particular, the last encoder block achieves the highest Fisher Ratio by a considerable margin, followed by the penultimate block, indicating that semantic class separation becomes highly concentrated near the output of the network rather than being uniformly distributed throughout the encoder.

\subsection{Class Centroid Distances}

\begin{figure}[H]
    \centering
    \includegraphics[width=\columnwidth]{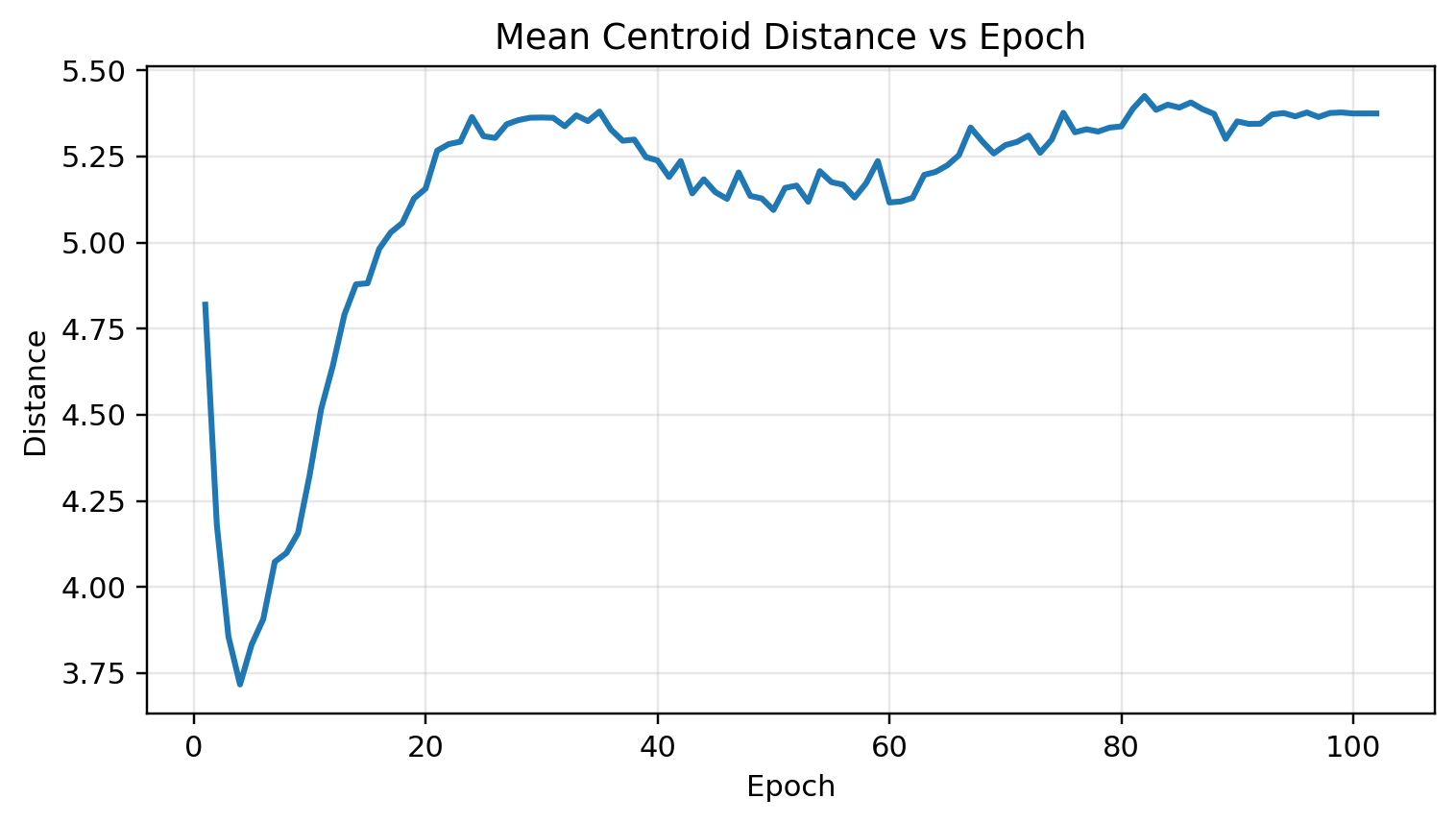}
    \caption{Evolution of the mean pairwise class centroid distance throughout Transformer training. The average Euclidean distance between all ImageNet-100 class centroids initially decreases during the early stages of optimization before increasing steadily as training progresses, indicating progressively improved global class separation within the learned representation space.}
    \label{fig:centroid_curve}
\end{figure}

To investigate the global organization of semantic classes, pairwise Euclidean distances between class centroids are computed throughout training. Figure~\ref{fig:centroid_curve} illustrates the evolution of the mean centroid distance. The mean centroid distance initially decreases during the first few epochs before increasing rapidly throughout optimization. After approximately twenty epochs, the centroid distances stabilize at substantially larger values than those observed during initialization and remain relatively constant for the remainder of training. The progressive increase in centroid separation indicates that semantic classes occupy increasingly distant regions within the learned representation space as optimization proceeds. This observation demonstrates a continuous expansion of the global semantic geometry throughout training.

\subsection{Local PCA Rank}

To quantify the local geometric complexity of semantic classes, Principal Component Analysis is performed independently for every ImageNet-100 class. Figure~\ref{fig:local_pca_curve} illustrates the evolution of the mean Local PCA Rank throughout optimization, while Figure~\ref{fig:local_pca_final} presents the class-wise Local PCA Rank at convergence.

Unlike the global semantic metrics, the Local PCA Rank exhibits a markedly different behaviour. During the early stages of optimization, the average Local PCA Rank increases rapidly, indicating that local semantic manifolds initially expand into higher-dimensional subspaces. After reaching a maximum during the first quarter of training, the Local PCA Rank decreases steadily until convergence. This gradual reduction suggests that the initially expanded local manifolds become progressively more compact and concentrated as optimization proceeds. The class-wise analysis further reveals substantial variability across semantic categories. Different ImageNet-100 classes require different intrinsic numbers of principal components to explain their local variance, indicating that semantic categories possess inherently different geometric complexities. Some classes remain highly concentrated within relatively low-dimensional subspaces, whereas others occupy considerably richer local manifolds.
\subsection{Summary of Observations}

Collectively, the proposed observatories reveal a progressive evolution of semantic representation geometry throughout Transformer training. Linear Probe Accuracy consistently increases throughout optimization, indicating progressively improved linear separability of semantic classes. Class Centroid Distances similarly increase after an initial transient phase, demonstrating that semantic classes occupy increasingly separated regions of the representation space. The Fisher Ratio exhibits a two-stage evolution, initially decreasing before increasing steadily during later optimization, while layer-wise analysis reveals that the strongest discriminative specialization emerges within the final Transformer blocks. In contrast, Local PCA Rank follows a non-monotonic trajectory, initially increasing as local semantic manifolds expand before gradually decreasing toward convergence, suggesting increasing local concentration of class representations. Furthermore, class-wise Local PCA analysis reveals substantial variability in geometric complexity across different semantic categories. Together, these observations provide complementary evidence describing the progressive organization, separation, and specialization of semantic representations within the Transformer feature space.

\section{Hypotheses}
\label{sec:hyp}
The above section~\ref{sec:find} discussed at length, the observations that came out of TGO-III. This section dives deep into the projected discussion which branch out of the empirical results. Previous TGOs~\cite{tgoi}~\cite{tgoii} dived deep into spectral and representational dynamics of the transformer and came up with a few hypotheses to explain certain observed phenomena. These phenomena included Effective Rank and Entropy explosion as the training progressed, sudden decrease in SVCCA~\cite{svcca} and CKA~\cite{cka}, increase in Intrinsic Dimensions, and so on. while this observatory does not add to the hypotheses, it does provide enough empirical evidence to validate or rule out the previous theories.

\subsection{Hypothesis-I: Semantic Expansion Hypothesis}
TGO-II concluded with invalidating one of the Hypotheses of \emph{token diversification as a reason behind rank explosion}, but solidified the claims over semantic expansion. This is where TGO-III picks up, section~\ref{sec:frame} discusses how semantic expansions points towards the central agenda that the dimensional and directional expansions can be attributed to the encoding of various semantic concepts. The observations obtained throughout TGO-III provide consistent evidence supporting this hypothesis. Linear Probe Accuracy improves steadily throughout optimization, indicating that class identity becomes progressively easier to decode from the learned representations. Similarly, the Fisher Ratio exhibits a sustained increase during later stages of training, demonstrating that between-class variance increasingly dominates within-class variance. The mean pairwise centroid distance also increases after the initial optimization phase, suggesting that semantic classes become progressively separated within the representation manifold. Collectively, these observations indicate that the representational directions discovered during optimization are not allocated randomly, but instead organize the feature space according to semantic categories.

Interestingly, this semantic organization occurs simultaneously with the manifold expansion previously reported in TGO-II. As intrinsic dimensionality increases, the Transformer acquires additional representational degrees of freedom, while TGO-III demonstrates that these newly explored dimensions contribute to improved semantic discrimination rather than arbitrary geometric expansion. Consequently, the Effective Rank explosion observed in TGO-I can now be interpreted not merely as covariance expansion, but as covariance expansion driven by progressively richer semantic organization of the representation manifold. The class-wise Local PCA Rank further complements this interpretation. Although different ImageNet-100 categories occupy manifolds of varying intrinsic complexity, every class develops a stable low-dimensional local structure by the end of optimization. This suggests that semantic expansion is accompanied by local semantic concentration, where globally the representation manifold expands to accommodate increasing semantic diversity while locally each semantic category becomes increasingly compact and geometrically organized. Taken together, TGO-I demonstrated that the covariance spectrum expands throughout optimization, TGO-II showed that the underlying representation manifold simultaneously expands into higher intrinsic dimensions, and TGO-III establishes that these newly explored dimensions are increasingly utilized for semantic discrimination. Therefore, the collective evidence across all three observatories provides strong empirical support for the \textbf{Semantic Expansion Hypothesis}: the Effective Rank explosion observed during Transformer training is primarily driven by the progressive organization of the representation manifold into increasingly discriminative semantic directions rather than by token diversification or purely statistical covariance growth.

\subsection{Hypothesis II: Manifold Expansion Hypothesis}

TGO-II proposed the Manifold Expansion Hypothesis after observing the remarkable agreement between the evolution of TwoNN~\cite{twonn} intrinsic dimensionality and the covariance expansion previously reported in TGO-I. While this observation suggested that the increasing Effective Rank originated from the progressive expansion of the underlying representation manifold, TGO-II was unable to determine how these additional representational degrees of freedom were utilized. In particular, it remained unclear whether the expanding manifold simply increased representational capacity or whether it was organized toward a meaningful semantic objective. The observations obtained through TGO-III provide important evidence addressing this limitation. As demonstrated in Section~\ref{sec:find}, the progressive increase in Linear Probe Accuracy, Fisher Ratio, and Class Centroid Distances indicates that the additional representational capacity produced through manifold expansion is accompanied by increasingly discriminative semantic organization. Rather than exhibiting arbitrary geometric complexity, the expanded manifold becomes progressively structured according to semantic class identity. Simultaneously, Local PCA analysis reveals that although the global representation manifold continues to expand, individual semantic classes gradually become more locally concentrated during the latter stages of optimization. This indicates that manifold expansion and local semantic concentration occur simultaneously rather than representing contradictory phenomena. Taken together, the observations from TGO-I, TGO-II, and TGO-III suggest a coherent geometric interpretation of Transformer training. TGO-I established that the covariance spectrum progressively explores additional directions, TGO-II demonstrated that these directions emerge through the expansion of the underlying representation manifold, while TGO-III shows that the newly available representational degrees of freedom are increasingly utilized to organize semantic classes into discriminative structures. Consequently, TGO-III significantly strengthens the Manifold Expansion Hypothesis by demonstrating that manifold expansion serves not only to increase representational capacity but also to facilitate progressively richer semantic organization throughout training.

\subsection{Hypothesis III: Transition Zone Hypothesis}

The Transition Zone Hypothesis was originally proposed in TGO-II after observing a consistent change in representational behaviour around the fourth and fifth Transformer blocks. Spectral observables measured in TGO-I exhibited remarkably similar dynamics within the middle Transformer layers, while TGO-II further demonstrated that representational similarity decreased and intrinsic dimensionality continued to increase beyond this region. These observations suggested the existence of a transition separating two qualitatively different stages of representation processing.

TGO-III introduces complementary semantic observatories capable of further examining this hypothesis. Layer-wise Linear Probe Accuracy demonstrates that semantic discrimination is not uniformly distributed throughout the Transformer depth. Instead, representations become progressively more linearly separable as depth increases, with the strongest discriminative capability emerging within the final Transformer blocks. Similarly, the layer-wise Fisher Ratio exhibits a highly non-uniform distribution, remaining relatively stable throughout most intermediate layers before increasing sharply within the final encoder blocks immediately preceding the classification head. These observations suggest that semantic specialization develops progressively throughout the network rather than appearing uniformly across all layers. While the representational transition originally identified in TGO-II remains evident from the perspective of geometry and intrinsic dimensionality, TGO-III indicates that the strongest semantic discrimination emerges substantially later within the network. Consequently, the Transition Zone should not necessarily be interpreted as the location where semantic representations become fully established. Instead, it is more appropriately viewed as the onset of geometric specialization, after which semantic organization continues to accumulate and mature throughout the remaining Transformer layers. Therefore, TGO-III refines rather than replaces the original Transition Zone Hypothesis. The collective observations suggest that Transformer training consists of multiple stages of representational evolution, beginning with geometric reorganization in the intermediate layers and culminating in highly discriminative semantic representations within the final encoder blocks.

\section{Conclusion}

This work introduced \textbf{TGO-III: Semantic Geometry Observatory}, the third observatory within the Transformer Geometry Observatory framework, with the objective of investigating how semantic organization emerges throughout Transformer training. Unlike the previous observatories, which primarily characterized global covariance evolution and representational geometry, TGO-III focused on the evolution of semantic class organization using Linear Probe Accuracy, Fisher Ratio, Class Centroid Distances, and Local PCA analysis. Collectively, these observatories enabled the direct investigation of class separability, local manifold geometry, and semantic representation throughout optimization. The observations obtained from TGO-III consistently demonstrate that semantic representations become progressively more discriminative as training proceeds. Linear Probe Accuracy increases steadily throughout optimization, indicating that class information becomes increasingly accessible through linear decision boundaries. Fisher Ratio analysis further reveals progressively stronger discrimination between semantic classes, while Class Centroid Distances demonstrate that semantic categories occupy increasingly separated regions within the representation manifold. Simultaneously, Local PCA analysis shows that although semantic classes exhibit varying geometric complexities, their local manifolds become progressively more concentrated during the latter stages of optimization.

These observations provide substantial empirical support for the \textbf{Semantic Expansion Hypothesis}. The covariance directions discovered during optimization are not merely additional statistical degrees of freedom but are increasingly allocated toward organizing semantic concepts. Consequently, the covariance rank expansion first observed in TGO-I can be interpreted as the geometric manifestation of progressively richer semantic organization rather than arbitrary covariance growth or token diversification. The present observatory also significantly strengthens the \textbf{Manifold Expansion Hypothesis}. TGO-I demonstrated that the covariance spectrum progressively expands throughout training, while TGO-II established that this expansion is accompanied by increasing intrinsic dimensionality of the underlying representation manifold. TGO-III completes this interpretation by demonstrating that the newly explored representational dimensions are increasingly utilized for semantic discrimination and class organization. Collectively, the three observatories suggest that manifold expansion provides the geometric substrate through which increasingly discriminative semantic representations emerge. Finally, TGO-III refines the \textbf{Transition Zone Hypothesis}. While previous observatories identified a geometric transition occurring within the middle Transformer layers through spectral and intrinsic-dimensional analyses, the present work demonstrates that semantic specialization continues to develop well beyond this region. Layer-wise Linear Probe Accuracy and Fisher Ratio reveal that the strongest discriminative representations emerge within the final Transformer blocks, suggesting that the intermediate transition corresponds to the onset of geometric specialization, whereas semantic discrimination progressively matures throughout the remaining depth of the network. Taken together, the three Transformer Geometry Observatories provide a unified geometric interpretation of Transformer learning. TGO-I established that the representation covariance progressively explores increasingly diverse directions, TGO-II demonstrated that these directions arise through the expansion of the underlying representation manifold, and TGO-III shows that this expanded manifold is progressively organized into increasingly discriminative semantic structures. This progression establishes a coherent picture of Transformer representation learning, connecting covariance evolution, manifold expansion, and semantic organization within a single geometric framework.

\section{Future Works}

TGO-II originally proposed six iterations of the \emph{Transformer Geometry Observatory}, where the remaining observatories were intended to investigate complementary aspects of Transformer representations. However, the observations presented throughout TGO-III substantially strengthen the semantic hypotheses proposed by the earlier observatories and provide a natural conclusion to the initial geometric investigation. This also brings us to a another point, that we have made clear progress in order to determine how a transformer may be learning. There are a further theories that need to be analyzed and tested. Consequently, the future trajectory of the Transformer Geometry Observatory has been slightly modified. Rather than extending semantic analysis further, the remaining observatories will investigate the computational and optimization mechanisms responsible for the spectral, representational, and semantic behaviours established throughout TGO-I, TGO-II, and TGO-III. Tracking Trajectories, understanding computational geometry and analyzing loss landscapes and how learning prog`resses.

\subsection{TGO-IV: Developmental Topology Observatory}

While TGO-III demonstrates the emergence of progressively richer semantic organization, it does not directly explain how Transformer computations produce these representations. Future work will therefore investigate the joint evolution of token dynamics and attention geometry through token trajectories, token interactions, attention routing, information flow, and head specialization. These experiments aim to determine how the underlying computational dynamics progressively construct the semantic manifolds observed throughout training.

\subsection{TGO-V: Gradient Geometry Observatory}

Having established the computational evolution of Transformer representations, future work will investigate the geometry of gradient propagation throughout optimization. TGO-V will analyze gradient covariance, gradient similarity, Fisher Information, gradient phase transitions, and layer-wise learning dynamics. These experiments aim to determine how gradient evolution governs the progressive emergence of structured Transformer representations.

\subsection{TGO-VI: Optimization Dynamics Observatory}

Future work will investigate the optimization landscape governing Transformer learning through Hessian spectra, curvature evolution, loss landscape topology, optimization trajectories, sharpness, and optimization phase transitions. These experiments seek to establish the optimization principles responsible for the geometric and semantic evolution consistently observed throughout the previous Transformer Geometry Observatories.

\subsection{Long-Term Objective}

The long-term objective of the Transformer Geometry Observatory is to establish a unified mechanistic theory describing how Transformers learn. By integrating spectral geometry, representation geometry, semantic geometry, computational dynamics, gradient geometry, and optimization dynamics, the complete framework aims to transform empirical geometric observations into fundamental learning principles. Ultimately, these principles may provide the theoretical foundation required for the development of future neuromorphic learning systems and biologically inspired Transformer architectures.
\bibliographystyle{IEEEtran}
\bibliography{references}
\end{document}